\documentclass[letterpaper]{article} 
\ifdefined\pdfshellescape\else\newcount\pdfshellescape\fi 
\usepackage[preprint]{coevokg_arxiv}
\usepackage[hyphens]{url}  
\usepackage{graphicx} 
\graphicspath{{Figures/}} 
\usepackage{natbib}  
\usepackage{caption} 
\usepackage{algorithm}
\usepackage{algorithmic}
\usepackage{amsmath}
\usepackage{amssymb}
\usepackage{multirow}
\usepackage{booktabs} 
\usepackage{arydshln} 
\usepackage{xcolor}
\definecolor{gaincolor}{rgb}{0.0,0.55,0.0} 

\newcommand{\githublink}[1]{%
  \begingroup
  \leavevmode
  \pdfstartlink attr{/Border[0 0 0]} user{/Subtype/Link/A<</S/URI/URI(#1)>>}%
  \textcolor{blue}{\url{#1}}%
  \pdfendlink
  \endgroup
}
\providecommand{\todo}[1]{--}
\newcommand{\gain}[1]{\textsubscript{\textcolor{gaincolor}{#1}}}
\newcommand{\colw}{4.6em}
\newcommand{\gv}[2]{\makebox[\colw][c]{#1\rlap{\,\gain{#2}}}}
\newcommand{\bv}[1]{\makebox[\colw][c]{#1}}
\newcommand{\hd}[1]{\makebox[\colw][c]{\textbf{#1}}}
\usepackage{newfloat}
\usepackage{listings}
\DeclareCaptionStyle{ruled}{labelfont=normalfont,labelsep=colon,strut=off} 
\floatstyle{ruled}
\newfloat{listing}{tb}{lst}{}
\floatname{listing}{Listing}
\title{CoEvoKG: Co-Evolving Knowledge Graphs with Self-Evolving Search Agents}
\author{
    Zhaoyang Li\textsuperscript{\rm 1}\equalcontrib,
    Zenghuang Fu\textsuperscript{\rm 2}\equalcontrib,
    Qiuyuan Ai\textsuperscript{\rm 1}\equalcontrib,\\
    Ping Jiang\textsuperscript{\rm 1,3},
    Haoyu Wu\textsuperscript{\rm 1,3},
    Minghui Wu\textsuperscript{\rm 1,3},
    Chenxu Zhao\textsuperscript{\rm 3},\\
    Jie Song\textsuperscript{\rm 1}\corresponding,
    Guannan He\textsuperscript{\rm 1}\corresponding
}
\affiliations{
    \textsuperscript{\rm 1}Peking University\\
    \textsuperscript{\rm 2}Institute of Automation, Chinese Academy of Sciences\\
    \textsuperscript{\rm 3}Mininglamp Technology\\
    gnhe@pku.edu.cn
}

\begin{document}
\maketitle
\begin{abstract}
Large language models can improve with reinforcement learning for search agents, yet existing self play agents repeatedly generate tasks while discarding the knowledge gained during successful searches. We introduce \textbf{CoEvoKG}, a framework that turns a knowledge graph into both a source of verifiable training tasks and a persistent evidence memory for agent evolution. CoEvoKG jointly trains a task generator and a search agent: the generator creates multihop questions from entity chains sampled from the knowledge graph, while the agent learns from rewards for answer correctness and search trajectories whose entity paths are supported by graph evidence. When a search succeeds, CoEvoKG verifies and deduplicates the retrieved evidence, then writes it back to the corresponding graph nodes and edges. Future rounds reuse this enriched graph for task generation and reward computation, closing the loop between model self evolution and knowledge accumulation. Experiments on six QA benchmarks (NQ, TriviaQA, PopQA, HotpotQA, 2WikiMultiHopQA, and Bamboogle) with three backbone models show that CoEvoKG improves macro average accuracy over the corresponding base models by +11.2, +10.1, and +11.6 points on Qwen2.5-3B-Instruct, Qwen2.5-7B-Instruct, and Llama-3.1-8B-Instruct, respectively. Under matched training budgets, CoEvoKG further improves over competitive self play baselines and RL baselines for search agents by +2.6 to +3.7 macro average points across the three backbones. Code is available at \githublink{https://github.com/lazzy1225/CoEvoKG}.

\end{abstract}

\section{Introduction}

\begin{figure}[t]
\centering
\includegraphics[width=\columnwidth]{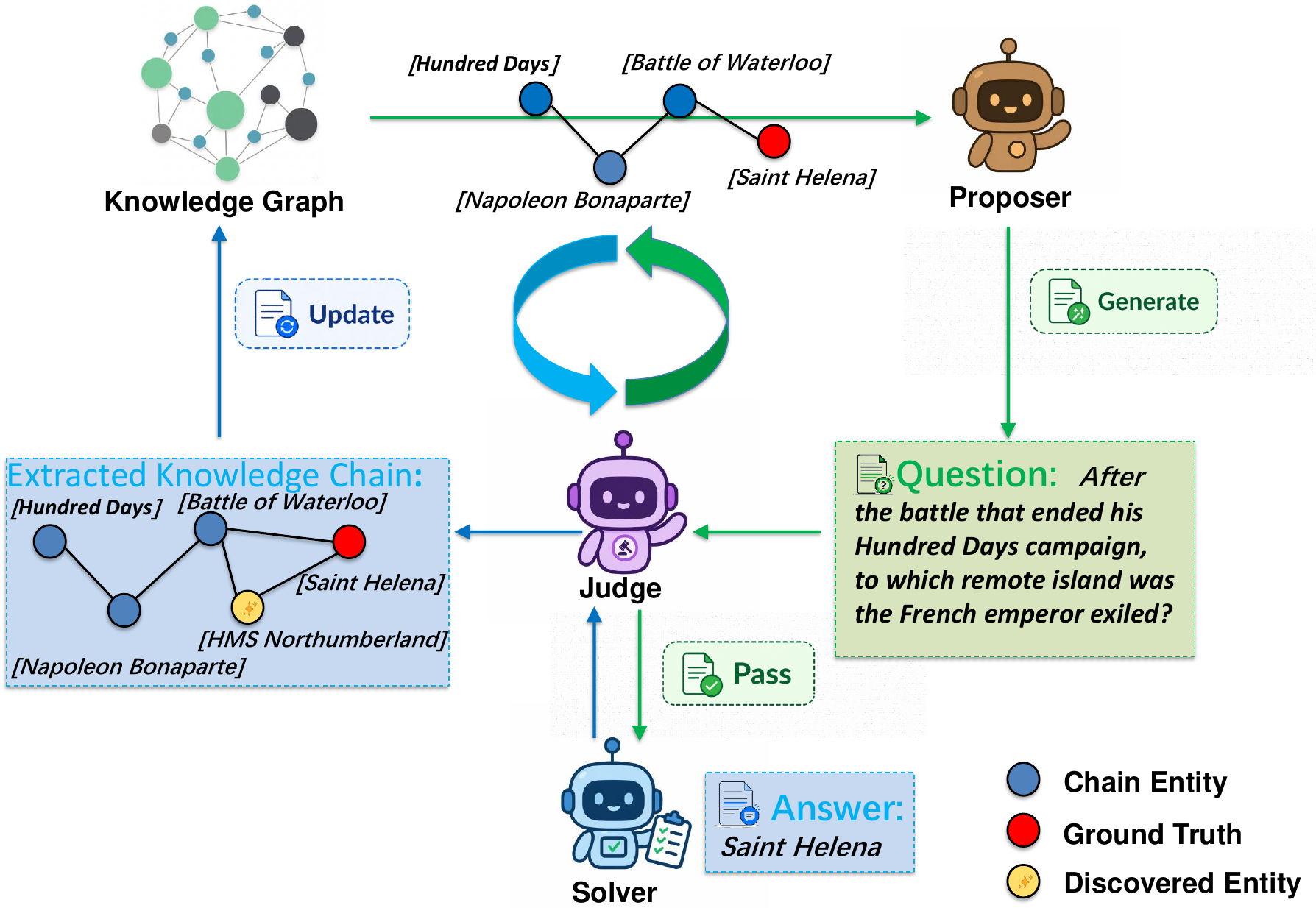}
\caption{CoEvoKG training loop. Multihop KG chains provide verifiable task templates and an external memory for solver evidence. Each iteration, the proposer generates questions from entity chains, the solver answers with multi turn search, and correct trajectories write verified evidence back to memory when their entity paths are supported by graph evidence.}
\label{fig:coevokg-overview}
\end{figure}

Reinforcement learning with verifiable rewards (RLVR) has become a powerful paradigm for improving the reasoning of large language models (LLMs)~\cite{guo2025deepseek,openai2024o1,shao2024deepseekmath}.
Extending it to \emph{agentic} settings, where a model iteratively invokes tools such as search engines to gather evidence across reasoning steps, is promising for knowledge intensive tasks~\cite{jin2025searchr1,chen2025research,zheng2025deepresearcher,zhang2026landscape}, but requires not only verifiable answer supervision but also a scalable supply of high quality search tasks and rewards over long trajectories.

Two obstacles stand in the way.
First, existing agentic RL methods train search using LLMs on fixed pools of human written QA~\cite{jin2025searchr1}, keeping reward verification tractable but bounding the training distribution; scaling to harder multihop tasks is costly, while unconstrained generation often yields ambiguous, unanswerable, or trivial questions.

Second, self evolution lets models generate and solve their own tasks~\cite{sukhbaatar2018selfplay,chen2024spin,lu2026search}, but the loop is unstable: generated tasks can be too easy, too hard, or unverifiable, with answer leakage or insufficient evidence. Moreover, once the solver is rewarded, the retrieved evidence is discarded, even though successful trajectories contain passages supporting the entity connections and answers.

To address these challenges, we propose \textbf{CoEvoKG} (Figure~\ref{fig:coevokg-overview}), a framework for self evolution in which knowledge graph (KG) entity chains play two roles: a structured source of verifiable multihop tasks, and an external memory enriched by solver verified evidence. Each step, a \emph{proposer} generates questions from multihop KG chains under a reward that targets a challenging but solvable difficulty band, while a \emph{solver} answers them with search and receives a reward combining answer correctness and path support; the evidence along correct trajectories is then written back (with deduplication) into the chains that generated it, so the proposer progressively samples from more informative contexts.

Empirically, on six QA benchmarks spanning general open domain QA (NQ~\cite{kwiatkowski2019nq}, TriviaQA~\cite{joshi2017triviaqa}, PopQA~\cite{mallen2023popqa}) and multihop reasoning (HotpotQA~\cite{yang2018hotpotqa}, 2WikiMultiHopQA~\cite{ho2020twowiki}, Bamboogle~\cite{press2023bamboogle}) across three backbones (Qwen2.5-3B-Instruct, Qwen2.5-7B-Instruct, and Llama-3.1-8B-Instruct), CoEvoKG improves accuracy over strong base models by \textbf{+11.2} (30.9$\to$42.1), \textbf{+10.1} (44.0$\to$54.1), and \textbf{+11.6} (39.8$\to$51.3) points.
More importantly, it surpasses representative self evolution and agentic RL search training methods (e.g., Search Self Play~\cite{lu2026search} and Search-R1~\cite{jin2025searchr1}) under matched training budgets, and we design ablations to separate the effects of our three contributions: task generation from KG chains, rewards for path support and difficulty, and evidence write back.

\begin{enumerate}
\item \textbf{Task generation from KG chains}: CoEvoKG uses multihop entity chains with relation labels to generate verifiable search questions, reducing reliance on human written QA and avoiding unconstrained question generation.
\item \textbf{Rewards for path support and difficulty}: the solver is rewarded for both answer correctness and path support in the graph, while the proposer is rewarded for producing questions near the solver's current difficulty frontier.
\item \textbf{Verified graph memory}: correct trajectories whose entity paths are supported by graph evidence are written back as reusable evidence, so later rounds use richer chains and close CoEvoKG's self evolution loop.
\end{enumerate}

\section{Related Work}

\begin{figure*}[t!]
\centering
\includegraphics[width=\textwidth]{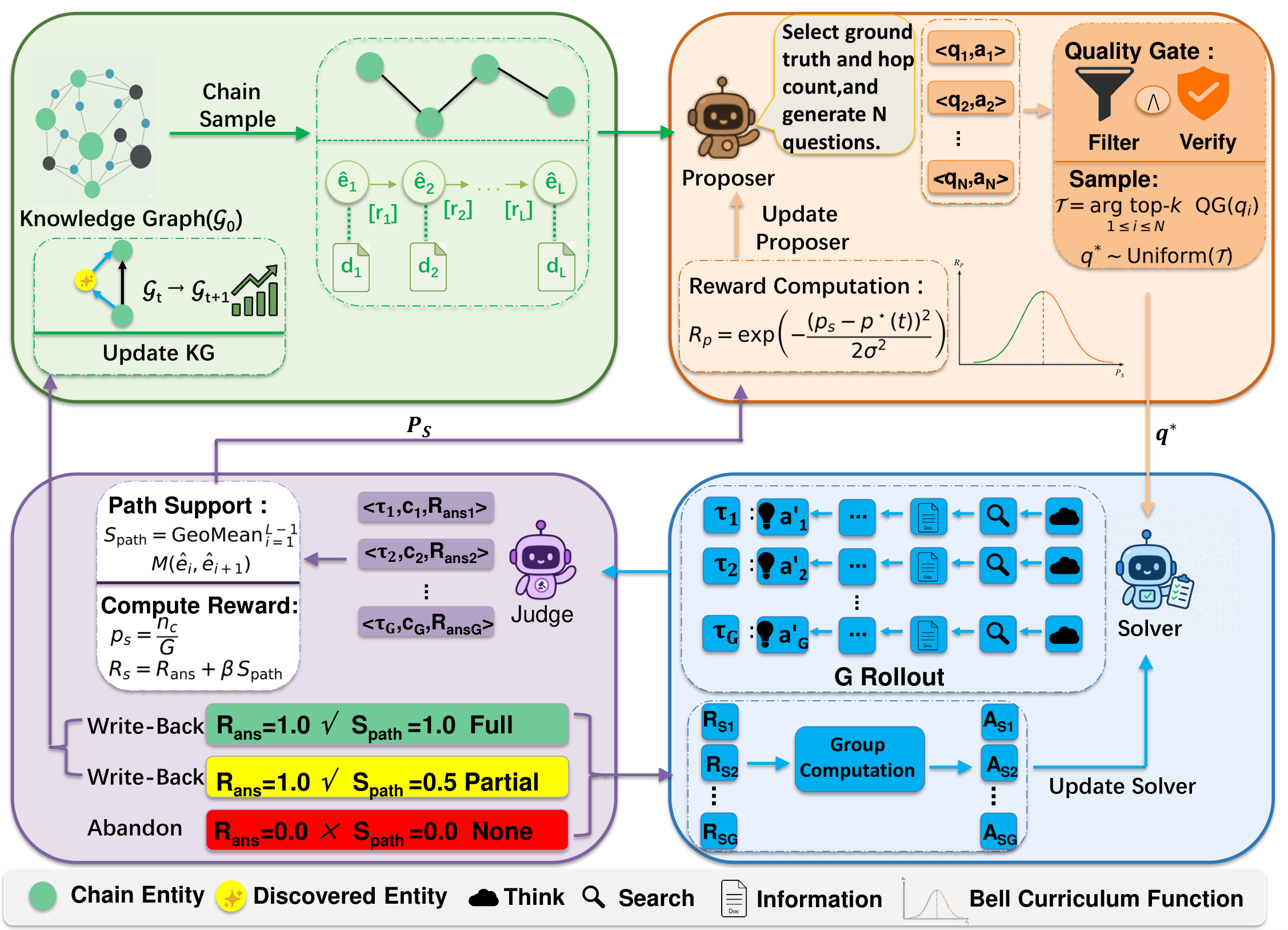}
\caption{CoEvoKG training loop. The proposer samples multihop entity chains from the current graph memory and generates candidate questions, which are filtered by a quality gate. The solver performs multi turn search rollouts and receives answer and path support rewards. Correct trajectories with path support are written back to the graph memory after verification and deduplication, enriching the chain pool for later proposer sampling.}
\label{fig:main}
\end{figure*}

\paragraph{RL for search agents.}
RL trains agents when and what to search: Search-R1~\cite{jin2025searchr1} and ReSearch~\cite{chen2025research} optimize multi turn trajectories with outcome rewards. Retrieval augmented generation grounds generation in retrieved evidence~\cite{lewis2020retrieval}, while Self-RAG learns when to retrieve and critique retrieved passages and generations~\cite{asai2024selfrag}; another line uses knowledge graphs at inference time to organize a corpus~\cite{edge2024graphrag} or walk the graph while reasoning~\cite{sun2024tog,long2025eperm}. These improve \emph{how} an agent solves fixed human written tasks; CoEvoKG instead treats the graph as a source of training tasks and a target for evidence accumulation.

\paragraph{Self improvement, generated tasks, and memory.}
Models increasingly generate their own training data via self play and self instruction~\cite{sukhbaatar2018selfplay,wang2023selfinstruct,chen2024spin,zhao2025absolute}. Closest to us, DeepDive~\cite{lu2025deepdiveadvancingdeepsearch} uses offline KG random walks to synthesize multihop questions for a fixed training set, while Search Self Play (SSP)~\cite{lu2026search} casts agentic search as a proposer--solver game but tends to degenerate and discards retrieved evidence. Built on RLVR~\cite{guo2025deepseek,openai2024o1} with GRPO~\cite{shao2024deepseekmath} for the solver and a REINFORCE++~\cite{hu2025reinforceplusplus} proposer, CoEvoKG keeps the KG in the training loop and, unlike episodic memory such as Reflexion~\cite{shinn2023reflexion}, writes verified entity evidence back into a graph store that is reusable as a task source. On the memory side, Cognitive Scaffold~\cite{ai2026cognitive} likewise externalizes a long context agent state into a persistent knowledge graph, crystallizing saturated context into structured snapshots for later retrieval; its graph, however, serves context management during inference, whereas CoEvoKG optimizes the graph inside the training loop and recycles it as verifiable multihop tasks. Concurrent self play work instead augments the RLVR signal itself to sustain diversity, e.g., variational problem synthesis~\cite{liang2026beyond}; CoEvoKG's gains stem from this persistent, verifiable evidence store and a process reward based on path support in the graph rather than from transient reward shaping over disposable episodes.

\section{Method}

CoEvoKG couples task generation and search solving around a shared pool of KG chains that both grounds generated tasks and stores evidence from successful solving. The design separates three concerns: how chains are sampled into questions, how solver trajectories are rewarded for answer and path support, and how verified evidence is merged back into memory. We first fix notation, then describe the three components in pipeline order: \emph{task generation from KG chains}, \emph{rewards for path support and difficulty}, and \emph{verified graph memory}, followed by optimization.

\subsection{Problem Formulation}

An evidence chain is a sequence of entity nodes linked by relations,
\begin{equation}
  c = (e_1, r_1, e_2, r_2, \ldots, r_{L-1}, e_L),
\end{equation}
where each node $e_i$ is a Wikipedia article paired with its passage $d_i$, and each $r_i$ is a typed natural language relation label (e.g., \emph{directed by}, \emph{located in}) linking $e_i$ to $e_{i+1}$. A chain thus couples a symbolic entity path with textual evidence at each node $\{d_i\}$, which the proposer reads to phrase questions and which later rounds enrich through evidence write back. Chains are drawn from a pool $\mathcal{G}$ populated by offline random walks over a Wikipedia based knowledge graph (the KILT knowledge source~\cite{petroni2021kilt}).

We use \emph{graph memory} to denote this evolving, entity indexed chain pool: it is induced by entity chains sampled from the underlying Wikipedia based knowledge graph, its nodes are the entities across all chains (each carrying its evidence passages) and its edges are the relations the chains realize, so chains sharing an entity become overlapping walks over a common graph, and it is updated online through verified evidence write back. We write $\mathcal{G}_t$ for the pool at round $t$ ($\mathcal{G}_0$ from offline walks); both path support verification and evidence write back operate on $\mathcal{G}_t$, the latter advancing it to $\mathcal{G}_{t+1}$.

Given a chain $c$, the proposer policy $\pi_p$ produces a question together with a target answer entity selected from the chain,
\begin{equation}
  (q, a) \sim \pi_p(\cdot \mid c), \qquad a \in \{e_1,\ldots,e_L\}.
\end{equation}
The solver policy $\pi_s$ answers $q$ by using search tool $\mathcal{T}$, producing a multi turn trajectory $\tau$ and final answer $y$,
\begin{equation}
  (\tau, y) \sim \pi_s(\cdot \mid q, \mathcal{T}).
\end{equation}
The two policies are optimized jointly under a self play objective,
\begin{align}
  \max_{\theta_s}\;& \mathbb{E}_{c\sim\mathcal{G},\,(q,a)\sim\pi_p,\,(\tau,y)\sim\pi_s}\!\left[R_s(q, a, c, \tau, y)\right],\\
  \max_{\theta_p}\;& \mathbb{E}_{c\sim\mathcal{G},\,(q,a)\sim\pi_p}\!\left[R_p(q, a, c; \pi_s)\right],
\end{align}
where the solver reward $R_s$ scores answer correctness together with the consistency between $\tau$ and the underlying chain, and the proposer reward $R_p$ scores the difficulty of admitted candidates relative to the current solver, while question quality is enforced separately by an upstream hard gate (Eq.~\ref{eq:quality}) rather than as a term in $R_p$. Crucially, $\mathcal{G}$ is not static. It is enriched online with evidence extracted from successful solver trajectories, so both expectations are taken over a distribution that coevolves with the policies.

\subsection{Task Generation from KG Chains}

This component realizes the first contribution: it turns the chain pool into a structured, verifiable task source for the proposer, reducing reliance on fixed human written QA without falling into the instability of unconstrained generation.

\paragraph{Evidence chain pool.}
Walking a large knowledge graph online is expensive, since each hop queries an external graph database. We hold a Wikipedia based knowledge graph (the KILT knowledge source), whose articles are linked by inline anchor hyperlinks, in MongoDB, and build the initial pool $\mathcal{G}_0$ offline by random walks over it: each $2$--$3$ hop walk yields a chain of article level entities, with each traversed edge annotated by a typed natural language relation label (e.g., \emph{directed by}, \emph{located in}), and each node keeping its article passage for question generation. We materialize these walks into the initial pool $\mathcal{G}_0$, sample directly from the current pool during training, and continually enrich it through evidence write back. We filter terminal entities that would surface in the question (list pages, bare years, very short tokens, nationality adjectives) and require at least two relational hops ($L\!-\!1\!\geq\!2$) to exclude single hop shortcuts. Full construction details are given in the appendix.

\paragraph{Chain based proposer.}
Conditioned on $c$ (the entity chain with its node passages $\{d_i\}$), the proposer selects an answer entity $a$ and emits one WH question that (i) names no chain entity verbatim, (ii) needs at least two hops, and (iii) has a unique, concise answer. Letting $a$ be any entity along the chain rather than the terminal one diversifies tasks and reduces positional leakage. We sample $M$ candidates per chain.

\paragraph{Quality gate.}
Generated questions are unreliable, so each candidate must pass a quality gate before training the solver. The gate is driven by an \emph{LLM verifier} that reads the question against its source chain and judges the properties a good multihop task needs: chain faithfulness, a genuine multihop requirement, a single answer focus, and clarity. A lightweight deterministic filter first removes broken candidates and answer leakage, and the verifier additionally enforces a hard no leakage constraint:
\begin{equation}
  Q(q,a,c) = \mathbb{1}[\text{no leak}]\cdot \mathrm{Verify}(q,a,c)\in[0,1].
  \label{eq:quality}
\end{equation}
Only candidates scoring above a threshold are retained; to keep verification cheap, all candidates of a chain are judged in a single grouped call. The verifier model, its prompt, the retention threshold, and how the dimension scores are aggregated are reported in the appendix.

\paragraph{Seed fallback.}
Because candidate generation is stochastic, some chains yield no question that passes the quality gate. Rather than leaving an empty solver batch slot, we fill it with a verified \emph{seed} question from a fixed human written QA set (the \emph{seed pool}). A seed slot uses the solver's normal reward machinery but provides no proposer difficulty signal, since no admissible generated question exists for that chain in the current round. This is purely a robustness device to keep batches well formed; its usage rate (the seed fallback rate) is monitored during training and declines from $\approx$0.40 to $\approx$0.25, remaining a minority throughout (Fig.~\ref{fig:curves}).

\subsection{Rewards for Path Support and Difficulty}

The second contribution is a single objective over long trajectories that couples the two policies through the evidence graph: the solver earns a path support reward when the searched entity path is supported by graph evidence, while the proposer earns a difficulty reward for placing questions at the solver's frontier.

\paragraph{Solver with search.}
The solver answers in multiple turns, interleaving reasoning with queries to a dense (E5) retriever~\cite{wang2022e5} over Wikipedia, so that training shapes \emph{search behavior} rather than answer generation alone. Outcome rewards alone, however, cannot tell reasoning supported by evidence from a lucky guess. We therefore pair the answer reward with a process reward that measures how well the trajectory is supported by the evidence graph:
\begin{equation}
  R_s = R_{\mathrm{ans}}\big(1 + \beta\,S_{\mathrm{path}}\big),
  \label{eq:solver-reward}
\end{equation}
where $R_{\mathrm{ans}}\!\in\!\{0,1\}$ is answer correctness, $S_{\mathrm{path}}\!\in\![0,1]$ is the path support score below, and $\beta$ weights the process term ($\beta{=}0.2$ in our runs). The process term is counted only for correct trajectories ($S_{\mathrm{path}}$ contributes only when $R_{\mathrm{ans}}{=}1$): among correct answers it separates reasoning supported by evidence from lucky matches, while an incorrect answer receives no process credit, so the reward never rewards an unsupported trajectory for merely surfacing plausible entities.

\paragraph{Path support verification.}
To judge whether the solver's search is genuinely supported by evidence, we first extract the entity path it traversed,
\begin{equation}
  \hat{c}_\tau = (\hat{e}_1, \hat{e}_2, \ldots, \hat{e}_L),
\end{equation}
determined jointly by the solver's reasoning steps, its queries, and the entity mentions in the retrieved passages. We verify $\hat{c}_\tau$ against the current evidence graph $\mathcal{G}_t$, whose nodes are entities and whose edges record verifiable associations between them together with the supporting passages.

Because the extracted entities are already grounded in the retrieved passages, verification reduces to checking, for each adjacent pair $(u,v)$, whether the trajectory's evidence attests a relation between them. We instantiate the relation consistency score $M(u,v)\in[0,1]$ deterministically without an LLM from the current chain evidence: over records mentioning both entities, a direct score $M_{\mathrm{direct}}(u,v)=\max_{d}\big(\alpha\,\mathrm{order}_d+\beta\,\mathrm{text}_d+\gamma\,\mathrm{co}_d\big)$ combines their ordering in the record's relation sequence ($\mathrm{order}$), their mutual mention in each other's passage ($\mathrm{text}$), and cooccurrence ($\mathrm{co}$); a one hop bridge $M_{\mathrm{bridge}}(u,v)=\eta\max_{z}\min\!\big(M_{\mathrm{direct}}(u,z),M_{\mathrm{direct}}(z,v)\big)$ captures indirect support, giving $M=\max(M_{\mathrm{direct}},M_{\mathrm{bridge}})$. The score is modular---any calibrated scorer (lexical, entailment, or learned) could replace it---and the weights $(\alpha,\beta,\gamma,\eta)$, candidate set, and thresholds are specified in the appendix.
The trajectory score is the geometric mean of the step supports,
\begin{equation}
  S_{\mathrm{path}} = \exp\!\Big(\tfrac{1}{L-1}\textstyle\sum_{i=1}^{L-1} \log\big(M(\hat{e}_i,\hat{e}_{i+1})+\epsilon\big)\Big),
\end{equation}
with $\epsilon$ a smoothing constant. The geometric mean is deliberate: if any single hop lacks support its score drops sharply, so a trajectory passes the path threshold ($S_{\mathrm{path}}\!\geq\!\tau_{\mathrm{path}}$) only when \emph{every} hop is traceable, not merely the endpoints. Used as the solver's process reward, $S_{\mathrm{path}}$ thus separates answers reached through evidence supported reasoning from lucky matches that are only correct at the final answer. The chain $\hat{c}_\tau$ is produced by the same LLM call that judges correctness, and the process reward is applied only to correct trajectories. Because each support value $M(\cdot,\cdot)$ is scored deterministically against the current evidence graph, independent of the answer, a correct trajectory earns a high $S_{\mathrm{path}}$ only when the entities it surfaces actually form a path supported by the graph---not simply because its final answer matches. On cheap string match cases, where no LLM extracted chain is available, the gold chain is used only to estimate task difficulty and as a prior for proposer scoring, never to grant solver process reward. The exact weights, the local support threshold, and the global threshold $\tau_{\mathrm{path}}$ are given in the appendix.

\paragraph{Difficulty reward for the proposer.}

The proposer is rewarded for placing questions at the solver's competence frontier, not for maximal hardness. For a candidate question we estimate its difficulty from the same group of $G$ solver rollouts used by GRPO: the success rate $p_s = n_c/G$ is the fraction of those rollouts whose \emph{final answer} is correct, where $n_c$ is the number of correct rollouts (we use final answer correctness, not the process augmented reward, so that partial path credit does not inflate $p_s$). We reward a target difficulty with a single peak bell reward whose target success rate is annealed over training,
\begin{equation}
  R_{\mathrm{diff}} = \exp\!\left(-\frac{(p_s - p^\star(t))^2}{2\sigma^2}\right),\quad
  p^\star(t) = p^\star_{s} - (p^\star_{s}-p^\star_{e})\tfrac{t}{T},
  \label{eq:diff}
\end{equation}
so questions that are trivially easy ($p_s\!\to\!1$) or unsolvable ($p_s\!\to\!0$) are discouraged while those landing near the current target $p^\star(t)$ are rewarded. The width $\sigma$ sets how tolerant the band is, trading selectivity for stability. Annealing the target from a higher initial value $p^\star_{s}$ to a lower final value $p^\star_{e}$ as the solver improves keeps generated questions challenging but solvable throughout training. We use the same schedule for all backbones (values in the appendix).

The proposer is then trained with this difficulty signal as its reward, $R_p = R_{\mathrm{diff}}$, while a candidate with failed extraction receives a small fixed penalty instead. Note that question quality $Q$ (Eq.~\ref{eq:quality}) acts upstream as a hard admission gate on candidates rather than as a term in $R_p$, so the proposer is optimized only over questions that already passed the quality bar. The schedule ($p^\star_{s}\!\to\!p^\star_{e}$, $\sigma$) and the group size $G$ are given in the appendix.

\subsection{Verified Graph Memory}

A key enabling component of CoEvoKG's self evolution loop is to preserve solver verified evidence as reusable graph memory. A correct trajectory has retrieved passages that connect entities and support the answer, yet this evidence is normally discarded once rewarded. CoEvoKG recycles it: by default, evidence write back is applied to trajectories that are both correct and supported by the path score ($R_{\mathrm{ans}}{=}1$ and $S_{\mathrm{path}}\!\geq\!\tau_{\mathrm{path}}$), attaching the retrieved entity evidence to the matching nodes and relations of the originating chain; requiring path support guards against admitting incidentally correct or weakly grounded trajectories. As each GRPO group yields only a few correct trajectories, preserving these scarce, already verified samples amortizes the cost of successful exploration. Evidence write back is deduplicated by a hash over the lowercased entity sequence: a new sequence becomes a new chain, a seen one is admitted again only with new evidence merged into its passages. Enriched chains return to the pool, forming a denser graph $\mathcal{G}_{t+1}$, so later rounds let the proposer condition on richer contexts and the verifier draw on more evidence.

\subsection{Optimization}

We instantiate both policies on a veRL~\cite{sheng2025hybridflow}\,+\,SGLang~\cite{zheng2024sglang} backend with asynchronous multi turn rollouts. Both maximize the same clipped PPO style policy gradient surrogate~\cite{schulman2017ppo} with a low variance ($k_3$) KL penalty to a frozen reference policy, and differ only in how the token level advantage $\hat{A}_t$ is formed.

\paragraph{Solver advantage (GRPO).}
The solver uses GRPO~\cite{shao2024deepseekmath}: for each question we sample $G$ trajectories and form group relative advantages, broadcast to every token of trajectory $i$,
\begin{equation}
  \hat{A}_{i,t} = \hat{A}_i = \frac{R_s^{(i)} - \mathrm{mean}(\{R_s^{(j)}\}_{j=1}^{G})}{\mathrm{std}(\{R_s^{(j)}\}_{j=1}^{G})+\epsilon},
  \label{eq:grpo-adv}
\end{equation}
removing the value critic while keeping stable credit assignment over long trajectories.

\paragraph{Proposer advantage (REINFORCE++).}
The proposer uses a REINFORCE++ style estimator~\cite{hu2025reinforceplusplus}: it forms cumulative discounted returns from the token rewards and whitens them across the global batch $\mathcal{B}$,
\begin{equation}
  G_t = \textstyle\sum_{s\geq t}\gamma^{\,s-t} r_s,
  \qquad
  \hat{A}_t = \frac{G_t - \mu_{\mathcal{B}}}{\sigma_{\mathcal{B}}+\epsilon},
  \label{eq:rpp-adv}
\end{equation}
with the difficulty reward $R_p$ received at the candidate's terminal token; normalization is global rather than group relative. As a lightweight stabilizer shared by both estimators, we additionally scale each generated token's advantage by an affine function of its own probability, damping updates on low probability tokens~\cite{yang2026do}. To keep the estimator on policy, each round resamples $M$ fresh candidates per chain; the quality gate (Eq.~\ref{eq:quality}) admits a top $k$ subset that is posed to the solver.

Each round proceeds in a fixed order: the proposer samples candidates from the current pool $\mathcal{G}_t$ and passes them through the quality gate; the solver then rolls out $G$ trajectories per admitted question, which yields both $R_s$ and, via the group success rate, the proposer signal $R_p$. We update both policies from this shared batch, then apply evidence write back \emph{after} the updates---advancing $\mathcal{G}_t$ to $\mathcal{G}_{t+1}$---so enriched chains take effect in the next round. The pool is refreshed continuously rather than on a separate schedule. Pseudocode is given in the appendix.

\begin{table*}[t]
\centering
\footnotesize
\setlength{\tabcolsep}{3pt}
\begin{tabular}{l c c c c c c c}
\toprule
\textbf{Model / Method} & \hd{NQ} & \hd{TriviaQA} & \hd{PopQA} & \hd{HotpotQA} & \hd{2Wiki} & \hd{Bamboogle} & \hd{Avg.} \\
\midrule
\textbf{Qwen2.5-3B-Instruct} & \bv{34.2} & \bv{52.0} & \bv{37.8} & \bv{30.6} & \bv{18.8} & \bv{12.0} & \bv{30.9} \\
\quad +Search-R1 & \gv{43.4}{+9.2} & \gv{54.4}{+2.4} & \gv{42.0}{+4.2} & \gv{32.4}{+1.8} & \gv{\textbf{31.6}}{+12.8} & \gv{26.4}{+14.4} & \gv{38.4}{+7.5} \\
\quad +SSP & \gv{38.6}{+4.4} & \gv{54.2}{+2.2} & \gv{39.2}{+1.4} & \gv{30.4}{-0.2} & \gv{25.4}{+6.6} & \gv{29.6}{+17.6} & \gv{36.2}{+5.3} \\
\quad \textbf{+CoEvoKG} & \gv{\textbf{46.4}}{+12.2} & \gv{\textbf{60.6}}{+8.6} & \gv{\textbf{45.6}}{+7.8} & \gv{\textbf{37.4}}{+6.8} & \gv{31.2}{+12.4} & \gv{\textbf{31.2}}{+19.2} & \gv{\textbf{42.1}}{+11.2} \\
\hdashline
\textbf{Qwen2.5-7B-Instruct} & \bv{43.0} & \bv{65.6} & \bv{45.0} & \bv{42.2} & \bv{35.4} & \bv{32.8} & \bv{44.0} \\
\quad +Search-R1 & \gv{55.2}{+12.2} & \gv{71.0}{+5.4} & \gv{42.6}{-2.4} & \gv{\textbf{55.6}}{+13.4} & \gv{38.4}{+3.0} & \gv{37.6}{+4.8} & \gv{50.1}{+6.1} \\
\quad +SSP & \gv{53.6}{+10.6} & \gv{74.4}{+8.8} & \gv{47.8}{+2.8} & \gv{49.4}{+7.2} & \gv{37.8}{+2.4} & \gv{42.4}{+9.6} & \gv{50.9}{+6.9} \\
\quad \textbf{+CoEvoKG} & \gv{\textbf{58.6}}{+15.6} & \gv{\textbf{75.4}}{+9.8} & \gv{\textbf{55.0}}{+10.0} & \gv{52.2}{+10.0} & \gv{\textbf{39.6}}{+4.2} & \gv{\textbf{44.0}}{+11.2} & \gv{\textbf{54.1}}{+10.1} \\
\hdashline
\textbf{Llama-3.1-8B-Instruct} & \bv{44.0} & \bv{65.0} & \bv{42.0} & \bv{36.2} & \bv{22.6} & \bv{28.8} & \bv{39.8} \\
\quad +Search-R1 & \gv{55.2}{+11.2} & \gv{67.2}{+2.2} & \gv{45.8}{+3.8} & \gv{43.2}{+7.0} & \gv{29.4}{+6.8} & \gv{43.2}{+14.4} & \gv{47.3}{+7.6} \\
\quad +SSP & \gv{54.6}{+10.6} & \gv{\textbf{75.2}}{+10.2} & \gv{48.2}{+6.2} & \gv{41.6}{+5.4} & \gv{30.8}{+8.2} & \gv{41.6}{+12.8} & \gv{48.7}{+8.9} \\
\quad \textbf{+CoEvoKG} & \gv{\textbf{57.2}}{+13.2} & \gv{74.4}{+9.4} & \gv{\textbf{50.8}}{+8.8} & \gv{\textbf{46.4}}{+10.2} & \gv{\textbf{33.6}}{+11.0} & \gv{\textbf{45.6}}{+16.8} & \gv{\textbf{51.3}}{+11.6} \\
\bottomrule
\end{tabular}
\caption{Main results: answer accuracy (\%) on six QA benchmarks across three backbones. Each block is headed by the backbone (instruction tuned base model); rows below it (\,+Search-R1, +SSP, +CoEvoKG) are search training methods applied to that backbone. Avg.\ is the macro average over the six benchmarks. Within each block, the best result in every column is in \textbf{bold}; subscripts give the absolute gain (points) over the backbone.}
\label{tab:main}
\end{table*}

\section{Experiments}

We evaluate CoEvoKG along three axes. First, we measure downstream QA accuracy against strong instruction tuned base models and against static data (Search-R1) and self play (SSP) RL baselines under a matched training budget, across six open domain benchmarks and three backbones. Second, we track the coevolution loop throughout training to verify that task generation and evidence memory improve together while the self play optimization stays stable. Finally, we attribute the gains to CoEvoKG's components through a contribution build up ablation and diagnose the quality of generated tasks.

\subsection{Experimental Setup}

\paragraph{Benchmarks.}
Following prior search agent work~\cite{jin2025searchr1}, we report on six open domain QA benchmarks: single hop NQ~\cite{kwiatkowski2019nq}, TriviaQA~\cite{joshi2017triviaqa}, and PopQA~\cite{mallen2023popqa}, and multihop HotpotQA~\cite{yang2018hotpotqa}, 2WikiMultiHopQA~\cite{ho2020twowiki}, and Bamboogle~\cite{press2023bamboogle}. The final evaluation uses the corresponding held out subsets from the public SSP release~\cite{lu2026search}: 500 questions per benchmark (125 for Bamboogle). These SSP evaluation subsets are used only for reporting; validation during training and seed fallback are split from the source benchmark training data and exclude the SSP evaluation examples. We report exact match accuracy after normalization and the macro average over the six.

\paragraph{Retrieval and interaction.}
All methods share one retrieval stack: a dense E5 retriever~\cite{wang2022e5} over a Wikipedia corpus, returning the top-3 passages per query. The solver runs for up to 8 turns, interleaving reasoning, search calls, and retrieved passages until it emits a final answer. We decode at temperature 0.6 with top-$p$ 0.95 during training, and use greedy decoding at evaluation for deterministic results.

\paragraph{Training.}
The solver is optimized with GRPO and the proposer with REINFORCE++ on a shared backbone in a self play loop, using a solver group size $G=8$ and an adaptive low variance KL penalty. The proposer constructs tasks by multihop walks over the chain pool, filtered by a quality gate, and evidence write back commits only solver trajectories with enough path support into the persistent graph memory. For a matched comparison, every method---CoEvoKG, Search-R1, SSP, and all ablations---is trained for the same fixed budget of $314$ optimization steps, and we select the checkpoint with the highest accuracy on a held out validation split drawn from training data; the selected checkpoint is then evaluated once on the SSP held out benchmark subsets. Full optimization, system, and data/threshold settings are given in the Technical Appendix.

\paragraph{Baselines.}
We evaluate three instruction tuned backbones spanning two model families and three scales---Qwen2.5-3B-Instruct, Qwen2.5-7B-Instruct, and Llama-3.1-8B-Instruct~\cite{qwen2025qwen25technicalreport,grattafiori2024llama3}---and compare three methods on each: (i) \textbf{Base}, the instruction tuned model with the same retrieval tool but no RL; (ii) \textbf{Search-R1}~\cite{jin2025searchr1}, agentic RL on a fixed pool of human written QA; and (iii) \textbf{SSP}~\cite{lu2026search}, proposer--solver self play without KG grounding or evidence write back.

\subsection{Main Results}

Table~\ref{tab:main} reports accuracy across the six benchmarks and three backbones. CoEvoKG improves the macro average over the base model by \textbf{+11.2} (30.9$\to$42.1) on Qwen2.5-3B-Instruct, \textbf{+10.1} (44.0$\to$54.1) on Qwen2.5-7B-Instruct, and \textbf{+11.6} (39.8$\to$51.3) on Llama-3.1-8B-Instruct, and attains the highest macro average of all methods under a matched training budget, also improving over the Search-R1 and SSP baselines. Gains are especially clear on several multihop and search intensive settings, while the strongest baseline remains competitive on a few individual datasets.

\subsection{Training Dynamics and Coevolution Signals}

\begin{figure*}[t]
\centering
\includegraphics[width=0.32\textwidth]{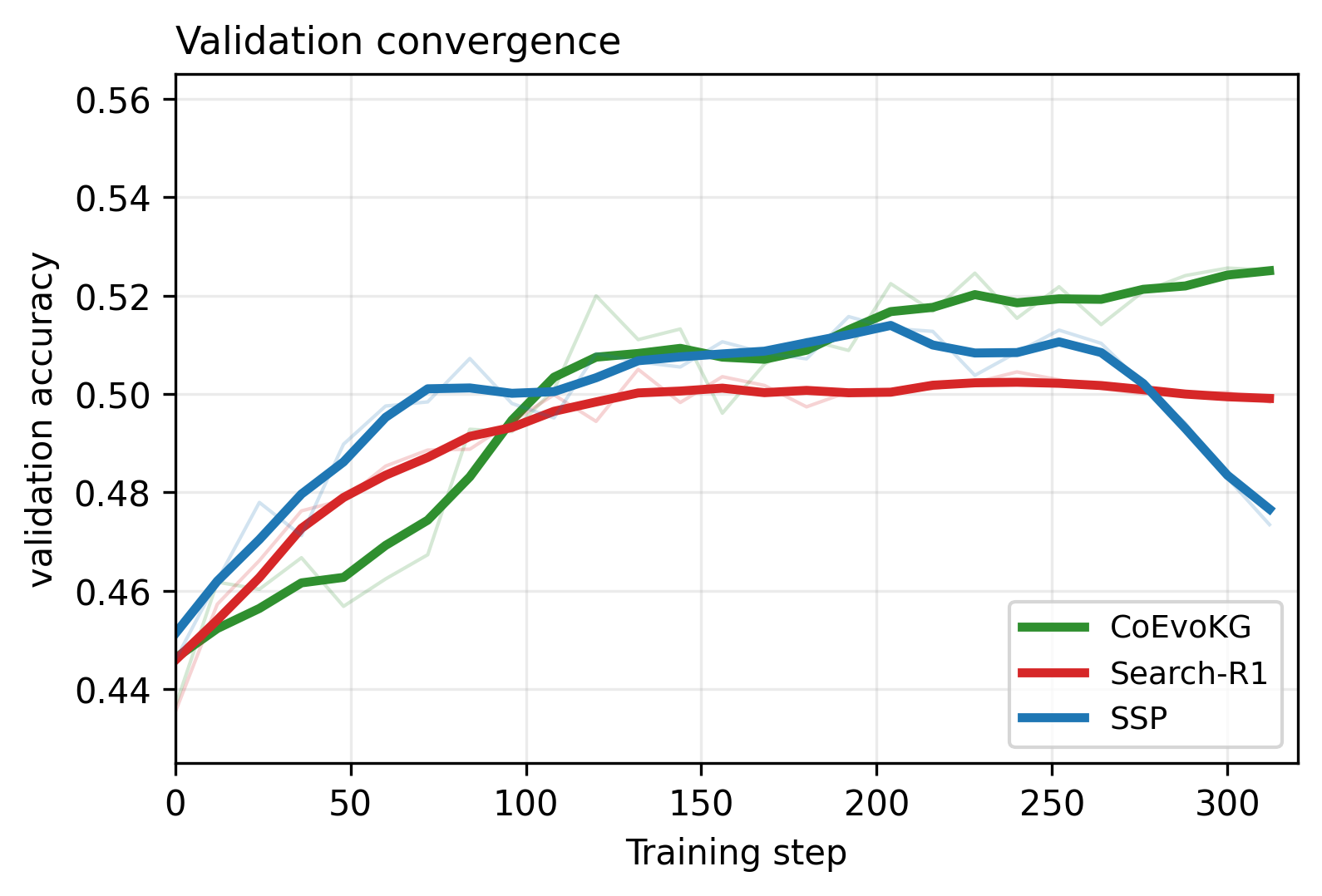}\hfill
\includegraphics[width=0.32\textwidth]{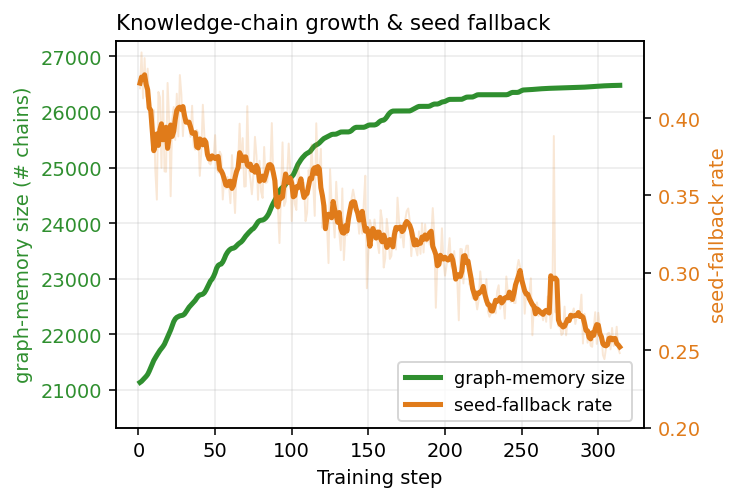}\hfill
\includegraphics[width=0.32\textwidth]{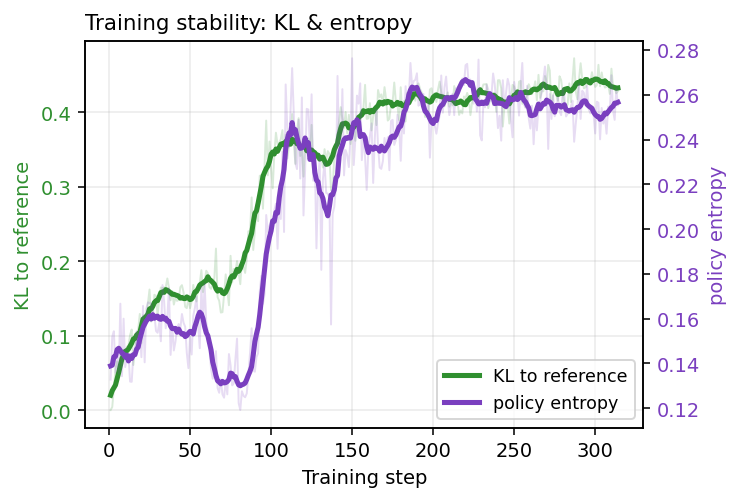}
\caption{Training dynamics on Qwen2.5-7B-Instruct (raw in light, running mean in bold). \textbf{(a)} Validation accuracy: CoEvoKG improves steadily under the matched budget and finishes highest, while SSP degrades after mid training and Search-R1 saturates early. \textbf{(b)} The verified chain pool grows then plateaus, and the seed fallback rate declines and stays a minority. \textbf{(c)} KL and policy entropy rise early then stabilize, without divergence or collapse.}
\label{fig:curves}
\end{figure*}

Figure~\ref{fig:curves} tracks the CoEvoKG run on Qwen2.5-7B-Instruct (our representative backbone) under the matched budget. CoEvoKG warms up more slowly than SSP in the first ${\sim}100$ steps but then keeps improving and finishes with the highest validation accuracy, whereas SSP degrades after mid training and Search-R1 saturates early---consistent with a coevolution loop that keeps supplying better grounded tasks rather than exhausting a fixed distribution. Meanwhile the verified chain pool grows then plateaus as deduplication admits fewer new chains, and the seed fallback rate declines and stays a minority, so the proposer keeps generating admissible questions rather than collapsing onto seeds. KL and policy entropy rise early and then stabilize, without divergence or entropy collapse.

\subsection{Generated Task Diagnostics}

\begin{figure}[t]
\centering
\includegraphics[width=\columnwidth]{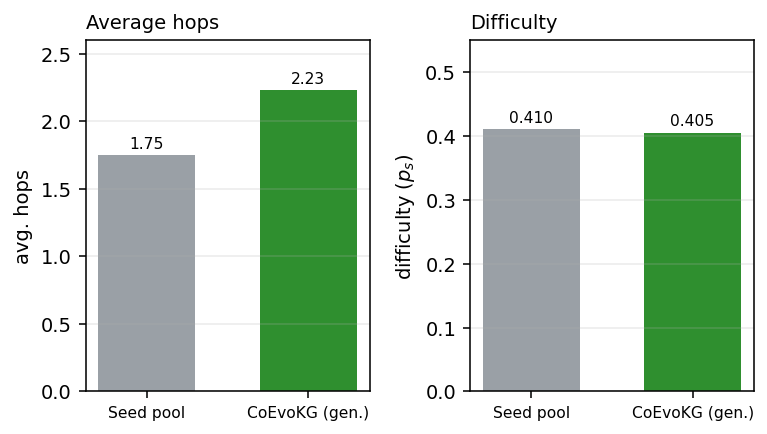}
\caption{Generated (CoEvoKG) vs.\ seed questions on Qwen2.5-7B-Instruct. Left: average number of hops (generated from the logged \texttt{avg\_selected\_hops}; seed estimated from its dataset composition). Right: difficulty as the mean solver success rate ($p_s$).}
\label{fig:qquality}
\end{figure}

Figure~\ref{fig:qquality} compares admitted generated questions with the seed pool on Qwen2.5-7B-Instruct. Generated questions have a comparable solver success rate ($p_s\!=\!0.405$ vs.\ $0.410$), suggesting the difficulty reward keeps them in a solvable range, while their average hop count is higher ($2.23$ vs.\ ${\approx}1.75$)---a stronger multihop bias consistent with CoEvoKG's larger gains on the multihop benchmarks (Table~\ref{tab:main}). This supports generation from KG chains as a source of nontrivial yet answerable search tasks.

\subsection{Ablations}

Because CoEvoKG forms a dependency stack, we use a build up ablation rather than leave one out: path support verification and evidence write back are only well defined once KG chains are used for task generation. Starting from SSP, we progressively add task generation from KG chains, rewards for path support and difficulty, and verified write back (Table~\ref{tab:ablation}), so the marginal gain of each row corresponds to one contribution. This build up also addresses reward hacking: the SSP starting point optimizes the adversarial $1{-}\mathrm{acc}$ objective most prone to unsolvable questions or answer leakage, and each added component---the no leakage quality gate (Eq.~\ref{eq:quality}), the difficulty reward, and the path support process reward---removes one such exploit. Because rows are scored by exact match on external, human curated benchmarks rather than the verifier used during training, pooled paired sign tests over the 2,625 held out examples complement the macro average deltas: C1/C2 are reliable, and write back is positive but weaker, with counts in the supplementary material.
\begin{table}[t]
\centering
\footnotesize
\setlength{\tabcolsep}{3pt}
\begin{tabular}{l c c c c c c}
\toprule
\textbf{Variant} & \textbf{C1} & \textbf{C2} & \textbf{C3} & \textbf{Avg.} & \textbf{$\Delta$} & \textbf{$p$} \\
\midrule
SSP (baseline)          & $\times$   & $\times$   & $\times$   & 50.9 & --- & --- \\
\;+ KG task generation  & \checkmark & $\times$   & $\times$   &  52.4 & \textbf{+1.5} & 0.0011 \\
\;+ path/difficulty rewards & \checkmark & \checkmark & $\times$   & 53.5 &  +1.1 & 0.0141 \\
\;+ write back (full)   & \checkmark & \checkmark & \checkmark & \textbf{54.1} &  +0.6 & 0.0314 \\
\bottomrule
\end{tabular}
\caption{Contribution build up ablation on Qwen2.5-7B-Instruct: starting from SSP, we add each component in dependency order. \textbf{C1}: task generation from KG chains; \textbf{C2}: rewards for path support and difficulty; \textbf{C3}: evidence write back. $\Delta$ is the macro average gain over the previous row; $p$ reports an exact two sided paired sign test against the previous row, with details in the supplementary material.}
\label{tab:ablation}
\end{table}

\section{Conclusion}
We presented CoEvoKG, a framework for self evolution where KG entity chains serve as both verifiable multihop task sources and persistent memory: the proposer generates chain grounded questions, the solver receives answer and path support rewards, and verified trajectories are written back for later rounds. Across six QA benchmarks and three backbones, CoEvoKG improves over strong base models and surpasses static data and self play RL baselines on macro average, with clear gains on several search intensive settings. Future work will automate evidence discovery, graph memory management, and extension beyond Wikipedia to scientific, legal, and biomedical domains.


\end{document}